\documentclass[a4paper, 10pt, twocolumn, twoside]{article}

\usepackage{ISARC}

\usepackage{lscape}
\usepackage{hologo}
\usepackage{amsmath}
\usepackage{bm}
\usepackage{booktabs}
\usepackage{makecell}
\usepackage{stfloats}

\begin{document}

\linespread{0.5}

\title{Terrain-Adaptive Mobile 3D Printing with Hierarchical Control}

\author{Shuangshan Nors Li$^{1}$ and J. Nathan Kutz$^{1,2}$}

\affiliation{
$^1$Department of Electrical and Computer Engineering, University of Washington, USA\\
$^2$Department of Applied Mathematics, University of Washington, USA
}

\email{
\href{mailto:ssli93@uw.edu}{ssli93@uw.edu},
\href{mailto:kutz@uw.edu}{kutz@uw.edu}
}

\maketitle

\thispagestyle{fancy}
\pagestyle{fancy}

\begin{abstract}
Mobile 3D printing on unstructured terrain remains challenging due to the conflict between platform mobility and deposition precision. Existing gantry-based systems achieve high accuracy but lack mobility, while mobile platforms struggle to maintain print quality on uneven ground. We present a framework that tightly integrates AI-driven disturbance prediction with multi-modal sensor fusion and hierarchical hardware control, forming a closed-loop perception-learning-actuation system. The AI module learns terrain-to-perturbation mappings from IMU, vision, and depth sensors, enabling proactive compensation rather than reactive correction. This intelligence is embedded into a three-layer control architecture: path planning, predictive chassis-manipulator coordination, and precision hardware execution. Through outdoor experiments on terrain with slopes and surface irregularities, we demonstrate sub-centimeter printing accuracy while maintaining full platform mobility. This AI-hardware integration establishes a practical foundation for autonomous construction in unstructured environments.
\end{abstract}


\section{Introduction}
\label{sec:Introduction}

Automated construction has long been a human aspiration \cite{you2025construction, bogue2017prospects}. It promises not only to accelerate labor-intensive tasks, but also to protect workers from danger by enabling construction in hazardous environments. Such capability is critical for establishing infrastructure in places that are otherwise inaccessible, including disaster recovery sites, remote locations, and challenging terrains.  Existing robotic systems have begun to address parts of this challenge. Gantry-based systems, such as contour crafting and large-scale 3D printing \cite{lim2012developments, gosselin2016large}, have demonstrated the ability to fabricate buildings with high geometric precision. However, significant challenges remain in material properties, geometric conformity, and process robustness \cite{buswell20183d}. Similarly, fixed robotic arms deployed on-site \cite{giftthaler2017mobile} provide flexible control and accurate material deposition. However, both approaches face fundamental limitations: gantry systems are immobile and require extensive setup, while fixed arms are restricted by their limited workspace. Both methods are therefore infrastructure-dependent and face challenges when applied to rapid deployment in unstructured environments.

Mobile robotic platforms offer a promising alternative by combining locomotion with manipulation capabilities \cite{keating2017toward, alhijaily2025development}. However, integrating high-fidelity construction with platform mobility remains a significant challenge \cite{dorfler2024advancing}: ground irregularities induce vibrations and pose disturbances that compromise print quality, while the dynamic nature of mobile bases conflicts with the precision requirements of material deposition.

\begin{figure*}[!htb]
    \centering
    \includegraphics[width=0.95\textwidth]{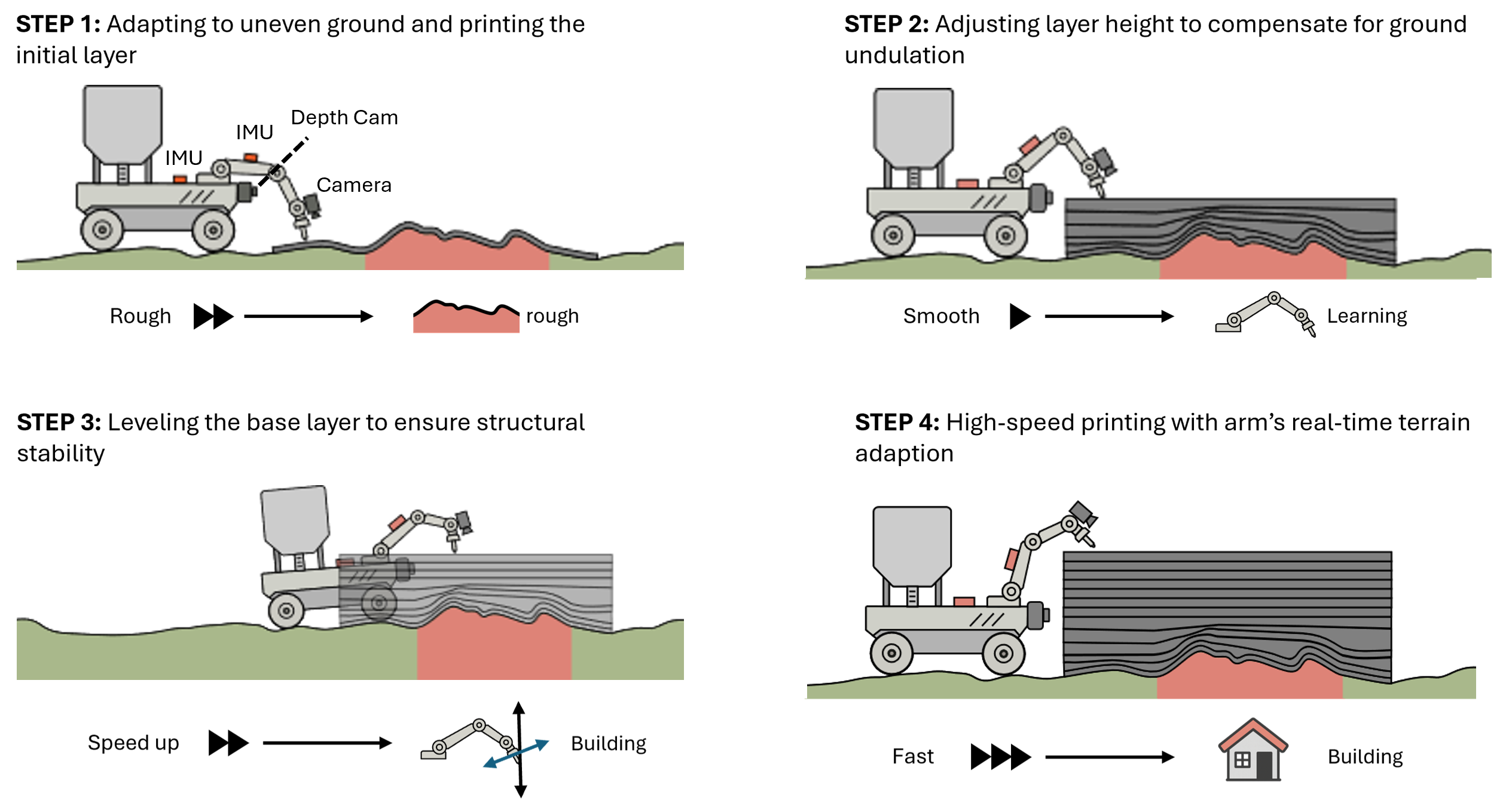}
    \vspace*{-.1in}
    \caption{Overview of terrain-adaptive mobile 3D printing. The mobile robot adapts to uneven ground and prints the initial layer (Step 1), adjusts layer height to compensate for terrain undulation (Step 2), levels the base layer to ensure structural stability (Step 3), and achieves high-speed printing with real-time terrain adaptation (Step 4).}
    \label{fig:overview}
\end{figure*}

Here we present a hierarchical control framework that addresses these challenges by integrating data-driven learning with precision hardware control (Figure~\ref{fig:overview}). The core idea is to leverage predictive models to learn environmental dynamics from sensor data and embed the learned knowledge into the hardware control loop, enabling intelligent real-time responses. Specifically, we developed a disturbance prediction module that learns terrain-to-perturbation mappings from multi-source sensors, predicting terrain-induced disturbances ahead of time to enable proactive compensation rather than reactive correction.

The main contributions of this work are:
\begin{itemize}[noitemsep]
    \item A three-layer hierarchical control architecture that separates path planning, predictive control, and hardware execution across appropriate time scales;
    \item Integration of data-driven disturbance prediction that enables proactive compensation for terrain-induced perturbations;
    \item Experimental validation demonstrating sub-centimeter printing accuracy on uneven terrain with full platform mobility.
\end{itemize}

\section{System framework}
\label{sec:Framework}

\begin{figure*}[t]
    \centering
    \includegraphics[width=0.9\textwidth]{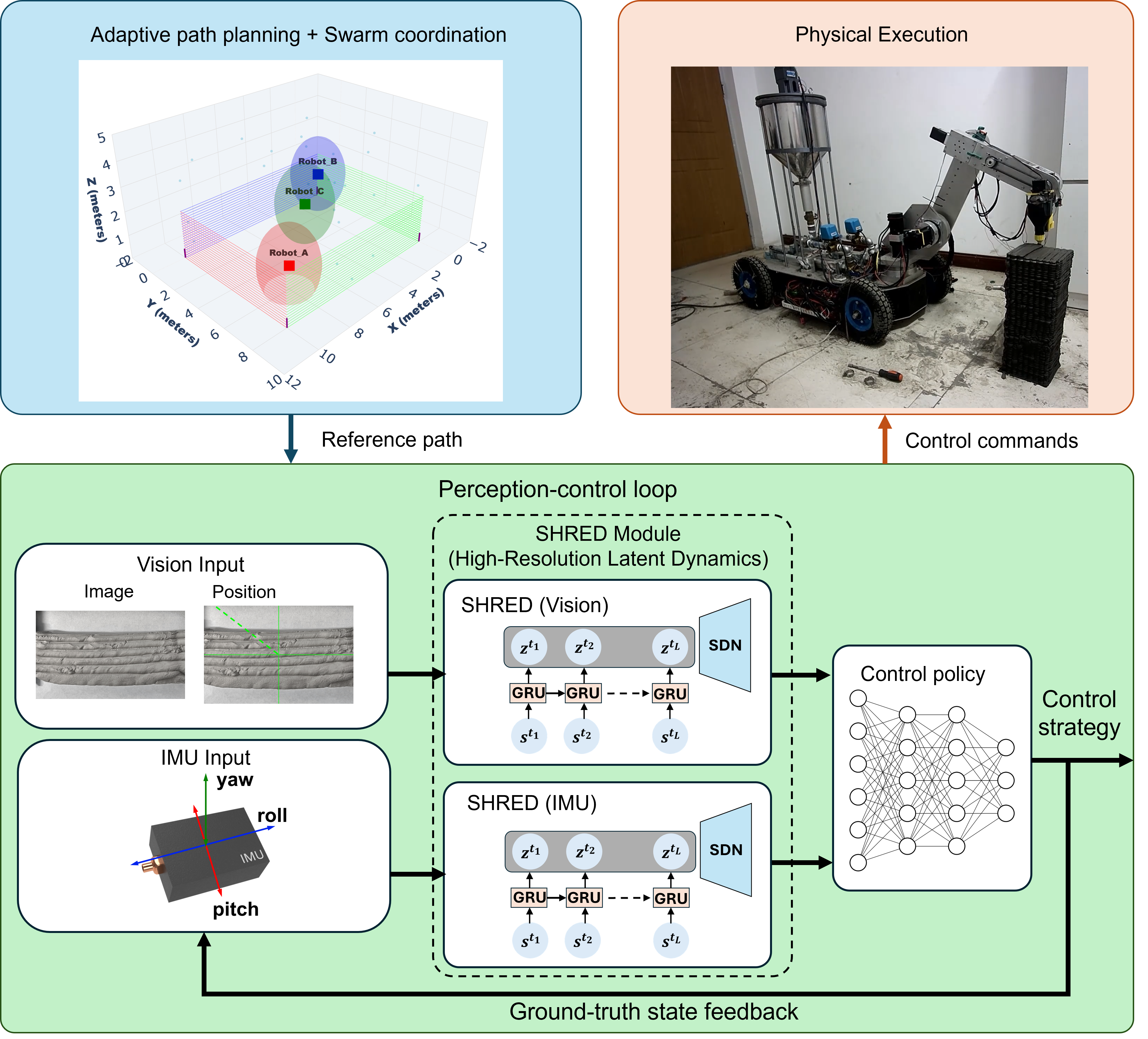}
    \vspace*{-.1in}
    \caption{Hierarchical control framework for terrain-adaptive mobile 3D printing. The system integrates adaptive path planning (left) with a perception-control loop (bottom). Sensor feedback from IMUs and vision-based tracking is processed through the disturbance prediction module, which enables feature selection for proactive compensation. The control policy generates real-time strategies for robotic actuation, enabling the mobile platform to maintain printing accuracy on unstructured terrain (right).}
    \label{fig:framework}
\end{figure*}

By integrating data-driven models with real-time sensor feedback and precision hardware control, we developed a terrain-adaptive printing framework that achieves both mobility and accuracy simultaneously (Figure~\ref{fig:framework}). The core philosophy is to enable the system to continuously learn environmental dynamics from sensor data and embed the learned predictive capabilities into the hardware control loop.

\subsection{Data-driven disturbance prediction}

The disturbance prediction module is the intelligent core of the framework. It learns terrain-to-perturbation mappings from multi-source sensors including IMU for chassis pose, RGB camera for end-effector tracking, and depth camera for terrain profiling. The module predicts 6-DOF disturbances $\hat{\bm{d}}_{t+\Delta t} \in \mathbb{R}^6$ (position and orientation perturbations) over a prediction horizon $\Delta t = 0.5$ s:
\begin{equation}
\hat{\bm{d}}_{t+\Delta t} = g(\bm{z}_t; \bm{\theta})
\label{eq:prediction}
\end{equation}
where $\bm{z}_t \in \mathbb{R}^{d}$ represents fused sensor features ($d$ ranges from 12 to 20 depending on sensor configuration) and $\bm{\theta}$ denotes learned model parameters. The input vector aggregates multi-modal information including IMU-derived linear acceleration and angular velocity, estimated base orientation, depth-derived terrain descriptors (local height variation and slope estimates), and end-effector pose relative to the mobile base. Raw images are not used directly; instead, compact geometric descriptors are extracted to enable real-time inference.

Sparse regression techniques \cite{kaiser2018sparse} are applied for feature selection, identifying the most predictive sensor channels for disturbance estimation. A lightweight feedforward network (two hidden layers with 128 and 64 units) processes the fused features and achieves inference latency of approximately 12 ms, enabling real-time proactive compensation rather than reactive correction \cite{silva2024real}.

The prediction module was trained on approximately 7 hours of operation data (4 hours from simulation, 3 hours from outdoor experiments) collected across four terrain types: flat concrete, grass-covered soil, loose gravel, and mixed terrain with slopes. Data from both sources were combined directly for training; the model's ability to generalize across these domains was validated through the outdoor experimental results. The dataset comprises approximately 1,400 trajectory segments totaling over 200 meters of traversed path, sampled at 50 Hz. Training and validation sets were split 80/20 at the trajectory level to avoid temporal correlation. Cross-time generalization was validated by resuming a printing task after a 24-hour interruption; the model maintained consistent prediction performance, demonstrating robustness to temporal discontinuities encountered in practical construction scenarios.

\subsection{Three-layer hierarchical architecture}

This prediction module is deeply integrated with the hierarchical hardware control architecture, forming a complete perception-learning-actuation closed loop:

\textbf{Upper layer (0.1 Hz)} handles adaptive path segmentation and velocity planning, transforming time-based printing trajectories into path-parameterized representations for speed-independent precision control. The path parameterization maps the desired trajectory $\bm{p}(t)$ to a parameter-based form:
\begin{equation}
\bm{p}(s) = (1-s)\bm{p}_0 + s\bm{p}_f, \quad s \in [0,1]
\label{eq:path}
\end{equation}
where $\bm{p}_0$ and $\bm{p}_f$ denote initial and final positions, enabling velocity-independent trajectory tracking. Complex trajectories are decomposed into piecewise-linear segments at this layer. This layer dynamically adjusts path complexity based on geometric features and terrain conditions, ensuring smooth transitions between printing segments.

\textbf{Middle layer (10 Hz)} implements preview model predictive control (MPC), receiving disturbance predictions and coordinating chassis-manipulator motion \cite{katayama2023model, stamatopoulos2025fully}. At each control cycle, the MPC solves:
\begin{equation}
\min_{\bm{u}} \sum_{k=0}^{N} \left( \|\bm{e}_k\|^2_{\bm{Q}} + \|\Delta\bm{u}_k\|^2_{\bm{R}} \right)
\label{eq:mpc}
\end{equation}
subject to system dynamics $\bm{x}_{k+1} = f(\bm{x}_k, \bm{u}_k, \hat{\bm{d}}_k)$ and operational constraints (joint limits, velocity bounds), where $\bm{e}_k$ is tracking error, $\bm{u}_k$ is control input, and $\hat{\bm{d}}_k$ is the predicted disturbance from Eq.~\eqref{eq:prediction}. The key innovation is a frequency-decomposition coordination strategy: the chassis handles low-frequency ($<$1 Hz), large-amplitude motions for terrain following, while the manipulator provides high-frequency ($>$5 Hz) precision corrections for accurate deposition.

\textbf{Lower layer (100 Hz)} provides precision hardware execution, controlling chassis steering, wheel speeds, and joint torques. High-frequency operation ensures real-time execution of MPC commands, while sensor data is fed back to the prediction module for continuous learning and model updates.

This hierarchical architecture enables the system to develop adaptive learning capabilities: as the robot operates across different terrains, the prediction module continuously accumulates experience and improves prediction accuracy.

\section{Validation}
\label{sec:Validation}

\subsection{Simulation validation}

\begin{figure}[!htb]
    \centering
    \includegraphics[width=0.48\textwidth]{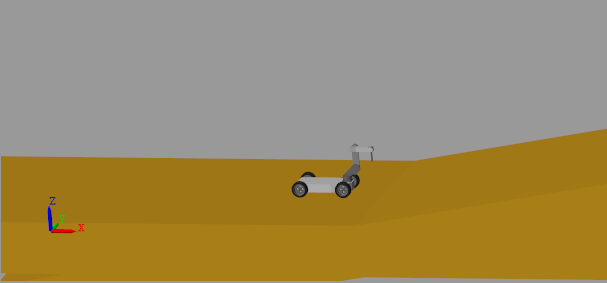}
    \caption{Simulink simulation environment showing the mobile robot traversing a slope terrain. The trajectory includes a 5-meter flat section followed by a 5-meter slope at 5° inclination.}
    \label{fig:sim_env}
\end{figure}

To validate the hierarchical control framework before physical deployment, we conducted dynamics simulations in MATLAB/Simulink. The simulation environment modeled a 10-meter trajectory comprising a 5-meter flat section followed by a 5-meter slope at 5° inclination (Figure~\ref{fig:sim_env}). To emulate terrain-induced disturbances, Gaussian noise with $\pm$5 mm amplitude was injected into the wheel transformation matrices (4$\times$4 homogeneous transforms containing orientation and position). While real terrain disturbances exhibit spatially correlated, non-Gaussian characteristics, the simulation provides a controlled environment to verify the MPC's disturbance rejection capability; the outdoor experiments subsequently confirm generalization to realistic disturbance patterns.

The MPC controller operated with a prediction horizon of 10 steps and control horizon of 5 steps, running at the middle control layer (10 Hz). Simulations were executed with a fixed-step solver at 1 ms resolution over a 30-second trajectory, representing continuous mobile printing operation.

\begin{figure}[!htb]
    \centering
    \includegraphics[width=0.48\textwidth]{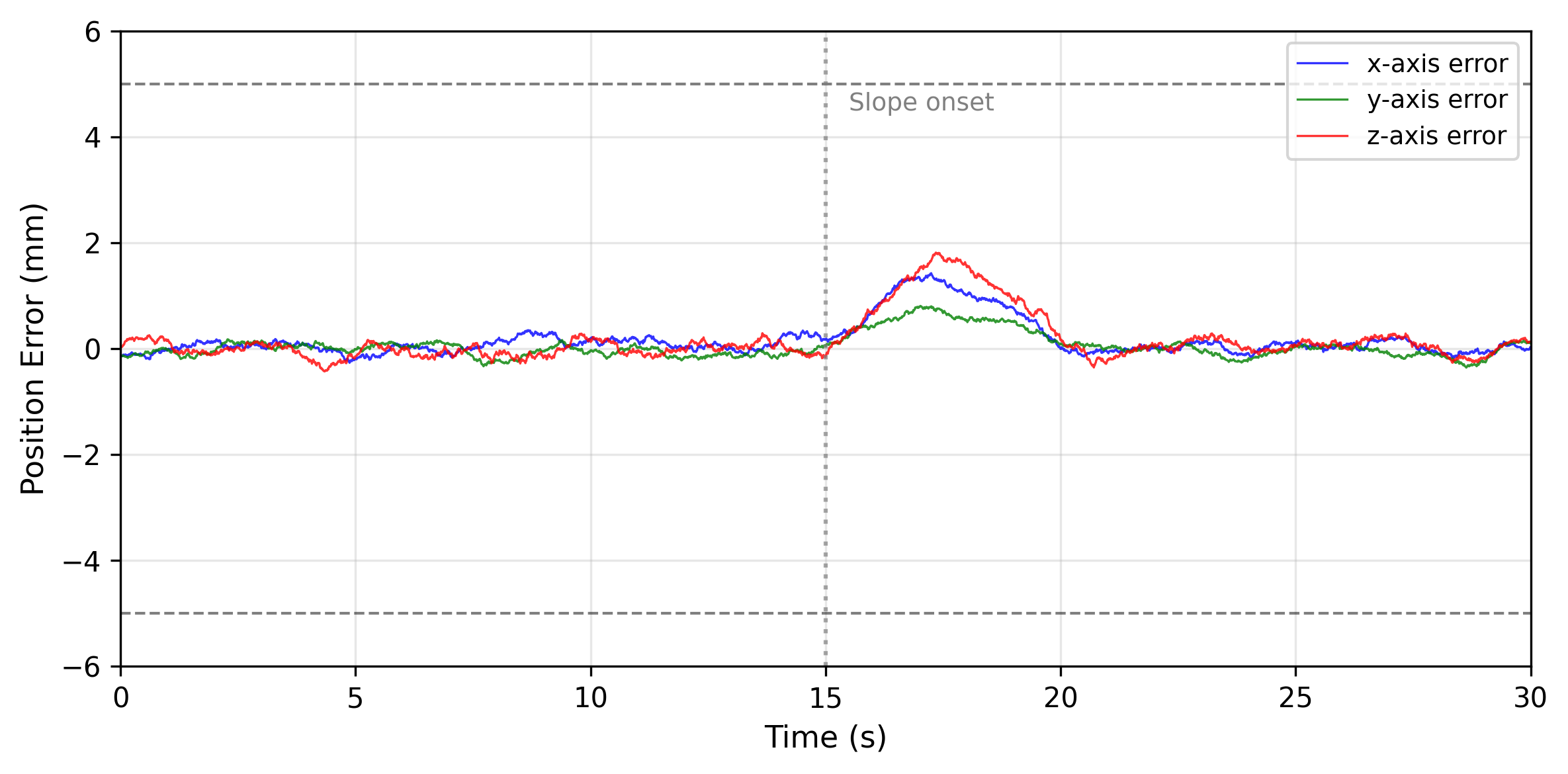}
    \caption{Simulated end-effector position error over a 30-second trajectory. Despite continuous terrain disturbances from slope transition (dashed line) and injected noise, the MPC-based compensation maintains position errors below 5 mm in all three axes.}
    \label{fig:simulation}
\end{figure}

Figure~\ref{fig:simulation} shows the end-effector position error throughout the simulated trajectory. Despite continuous terrain disturbances from both the slope transition and injected noise, the MPC-based compensation maintained position errors below 5 mm in all three axes. The error remained bounded without accumulation, validating the effectiveness of the predictive control strategy before field deployment.

\subsection{Experimental setup}

To further validate the terrain-adaptive capabilities on physical hardware, we conducted outdoor experiments on diverse terrain conditions. The experiments evaluated the system's ability to maintain printing precision despite ground irregularities.

\begin{figure}[!htb]
    \centering
    \includegraphics[width=0.48\textwidth]{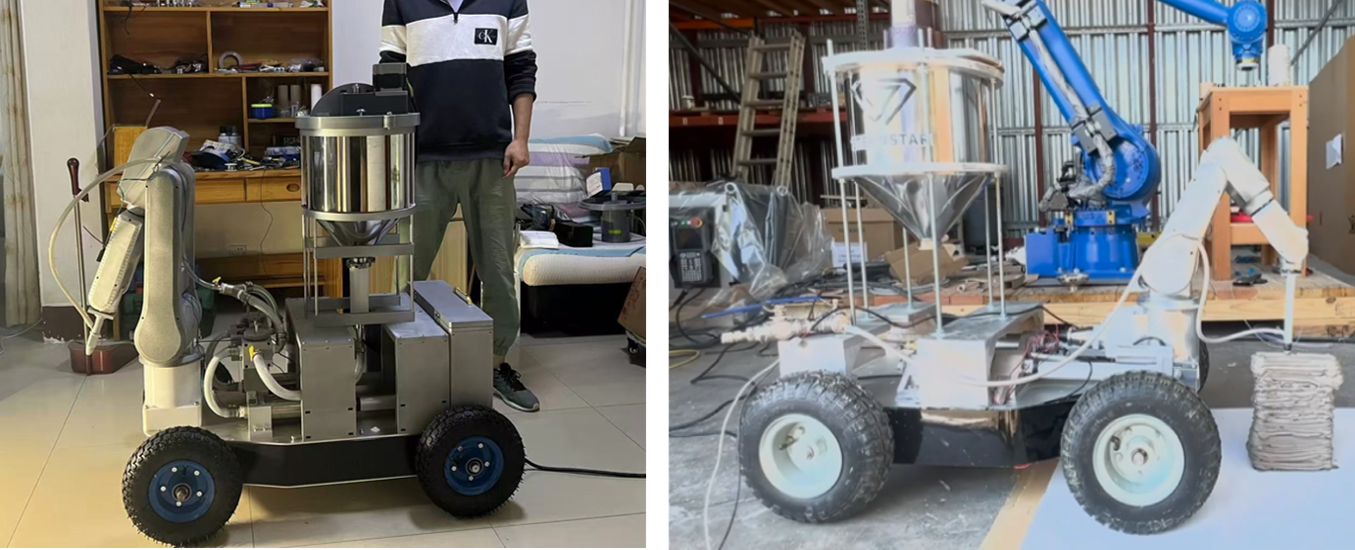}
    \caption{Experimental mobile 3D printing platform. Left: initial prototype with material hopper and extrusion system. Right: integrated system with 6-DOF robotic arm for terrain-adaptive printing.}
    \label{fig:platform}
\end{figure}

The experimental platform consisted of a four-wheel drive mobile base (1.2 m $\times$ 0.8 m, 150 kg payload capacity) equipped with a 6-DOF robotic arm (1.5 m reach, 10 kg payload), as shown in Figure~\ref{fig:platform}. Tests were conducted on terrain featuring slopes, surface roughness, and mixed surface types including grass, gravel, and concrete with irregularities. These conditions represent realistic challenges encountered in disaster recovery sites and undeveloped construction areas.

Sensing systems include: IMU (100 Hz) for chassis state estimation, joint encoders (1000 Hz) and force/torque sensors (500 Hz) for manipulator control, stereo vision (30 Hz) for end-effector tracking, and RGB-D camera (30 Hz) for terrain mapping.

\subsection{Performance results}

In the primary validation experiment, the mobile robotic platform constructed a concrete foundation structure on uneven ground. Although limited in physical size, this structure represents a fundamental construction primitive that can be repeatedly deployed to form larger foundations or continuous wall segments--in practice, large-scale construction can be decomposed into such locally adaptive printing units. Despite challenging ground conditions, the hierarchical control framework maintained fabrication accuracy within sub-centimeter level across the entire build, as confirmed by LiDAR scanning.

The disturbance prediction module proved critical for maintaining accuracy. By learning terrain-to-perturbation mappings, the system achieved effective disturbance rejection across different terrain types. Table~\ref{tab:terrain_accuracy} summarizes end-effector tracking accuracy across different terrain conditions.

\begin{table}[htb]
\centering
\caption{End-effector tracking accuracy during printing on different terrain types}
\label{tab:terrain_accuracy}
\begin{tabular}{@{}lccc@{}}
\toprule
Terrain & \makecell{Height Dev.\\(mm)} & \makecell{Misalign.\\($^\circ$)} & \makecell{Max Dev.\\(mm)} \\
\midrule
Flat concrete & 2.1 $\pm$ 0.4 & 0.08 $\pm$ 0.02 & 3.8 \\
Grass (soft) & 4.3 $\pm$ 1.2 & 0.21 $\pm$ 0.08 & 8.7 \\
Gravel & 5.1 $\pm$ 1.5 & 0.25 $\pm$ 0.09 & 9.2 \\
Slope (5$^\circ$) & 3.8 $\pm$ 0.9 & 0.18 $\pm$ 0.06 & 6.5 \\
Mixed & 4.7 $\pm$ 1.3 & 0.22 $\pm$ 0.07 & 8.9 \\
\bottomrule
\end{tabular}
\end{table}

Table~\ref{tab:system_metrics} presents additional system performance metrics, demonstrating real-time capability and rapid disturbance recovery.

\begin{table}[htb]
\centering
\caption{System performance metrics}
\label{tab:system_metrics}
\begin{tabular}{@{}lc@{}}
\toprule
Metric & Value \\
\midrule
MPC computation time & 8.3 $\pm$ 2.1 ms \\
Prediction latency & 12.5 $\pm$ 3.2 ms \\
Settling time & 0.35 $\pm$ 0.12 s \\
Layer height consistency & $\pm$0.8 mm \\
\bottomrule
\end{tabular}
\end{table}

The final structure exhibited consistent layer stacking and mechanical stability. Across all terrain types, the system maintained sub-centimeter accuracy (mean position error $<$ 5.1 mm), validating that the proposed framework enables mobile platforms to achieve precision approaching stationary gantry systems while operating on challenging terrain.

\section{Discussion and conclusion}
\label{sec:Conclusion}

This study presents a hierarchical control framework that addresses the fundamental challenge of maintaining deposition precision on mobile platforms operating over unstructured terrain. Unlike existing approaches that either sacrifice mobility for accuracy or accuracy for mobility, our three-layer architecture achieves both through temporal separation of control concerns and predictive disturbance compensation.

The key technical contributions include: (1) a three-layer hierarchical control architecture that separates path planning, predictive control, and hardware execution across appropriate time scales; (2) integration of data-driven disturbance prediction that enables proactive compensation for terrain-induced perturbations; and (3) dual-layer chassis-manipulator coordination through frequency decomposition.

\subsection{Failure modes and mitigation}

Mobile construction systems operating on unstructured terrain face several characteristic failure modes. Understanding these failure modes and how the proposed framework addresses them is essential for assessing deployment readiness.

\textit{Terrain perception degradation} occurs when environmental conditions compromise sensor accuracy. Varying lighting affects RGB-based end-effector tracking, while surface reflectivity and dust interfere with depth measurements. The multi-modal sensor fusion approach mitigates this by combining IMU, vision, and depth data, allowing the prediction module to maintain reasonable estimates even when individual sensors degrade. The learned terrain-to-perturbation mappings provide robustness by encoding statistical regularities rather than relying on instantaneous measurements alone.

Unlike laboratory robotic systems where disturbances are isolated events, construction sites exhibit persistent environmental uncertainty: airborne dust from excavation activities, rapidly changing shadows from moving equipment, and surface glare from wet concrete or reflective materials. These conditions demand sensor fusion strategies that degrade gracefully rather than fail catastrophically. The hierarchical architecture addresses this by allowing the prediction module to operate with reduced feature sets when specific sensors become unreliable, falling back to IMU-dominated estimation during severe visual degradation while maintaining acceptable---though reduced---compensation performance.

\textit{Wheel slip and chassis compliance} introduce discrepancies between commanded and actual platform motion, particularly on soft or loose surfaces. The frequency-decomposition coordination strategy addresses this by assigning terrain-following to the low-frequency chassis control loop while reserving the high-frequency manipulator for precision corrections. This separation ensures that chassis-level disturbances do not directly propagate to the deposition point.

\textit{Long-horizon drift} poses a risk for extended construction tasks where small errors could accumulate over time. The experimental results provide direct evidence against this concern: over a 5-hour continuous printing session comprising approximately 140 layers, no systematic error accumulation was observed. This stability arises from the closed-loop architecture where sensor feedback continuously updates the prediction module, preventing drift from compounding across layers.

Material rheology and extrusion dynamics can further influence deposition quality; however, this work focuses on terrain-induced kinematic disturbances, and material-process interactions will be addressed in future studies.

\subsection{Scalability implications}

The dominant barriers to scaling mobile construction systems are typically long-duration stability and site-to-site variability, rather than geometric size alone. The proposed framework addresses both concerns through its hierarchical design.

For duration scaling, the demonstrated 5-hour operation without error accumulation suggests that the architecture handles the primary failure modes that typically prevent mobile systems from sustaining construction-time-scale operations. The locally adaptive nature of the control framework---where each printing segment is treated as a path-parameterized unit with real-time disturbance compensation---means that scaling to larger structures involves repeating proven primitives rather than extrapolating untested behaviors.

For multi-robot scaling, the modular architecture provides a natural extension path. Each robot can operate its own perception-prediction-control loop while coordinating at the path-planning level. Recent work on aerial multi-robot construction \cite{zhang2022aerial} and collaborative mobile printing \cite{li2025frontiers} demonstrates the feasibility of such coordination, and the hierarchical separation in our framework aligns well with distributed control paradigms.

From a practical deployment perspective, the system requires a minimum sensor configuration of one IMU and either stereo vision or an RGB-D camera for basic operation; the full sensor suite enhances performance but is not strictly necessary for all terrain types. The framework assumes locally continuous terrain variations with spatial frequencies below approximately 0.5 cycles per meter---corresponding to undulations with wavelengths greater than 2 meters. Extremely discontinuous obstacles such as steps, curbs, or debris piles exceeding the manipulator's compensation range (approximately 15 cm vertical displacement) remain outside the current operational envelope and require path replanning at the upper control layer.

The construction workflow integration follows a modular primitive approach: rather than attempting continuous printing of entire structures, the system treats each foundation segment, wall section, or structural element as an independent printing unit with well-defined start and end conditions. This decomposition aligns naturally with construction sequencing practices, where curing time, material logistics, and inspection checkpoints create natural task boundaries. The demonstrated 5-hour operation capability corresponds to typical work shifts, suggesting compatibility with existing site management practices.

Several limitations remain before widespread deployment. The current system depends on accurate terrain mapping for optimal disturbance prediction, which may be challenging in rapidly changing environments such as active construction sites with concurrent human activity. Additionally, extending the framework to different construction materials will require characterizing material-specific disturbance responses. A systematic comparison with reactive-only control would further quantify the predictive module's contribution.

Looking ahead, the practical implications of this work extend to disaster recovery, remote infrastructure construction, and eventually to extraterrestrial construction where autonomous operation on irregular surfaces is essential. By demonstrating that precision and mobility can coexist in construction robotics, this framework establishes a foundation for autonomous building in environments previously inaccessible to conventional methods.

\bibliography{ISARC}

\end{document}